\documentclass{article}

\usepackage{microtype}
\usepackage{graphicx}
\usepackage{subfigure}
\usepackage{booktabs} 
\usepackage{multirow}
\usepackage{siunitx}
\usepackage{textcomp}
\usepackage{hyperref}

\usepackage[accepted]{icml2019}

\begin{document}

\twocolumn[
\icmltitle{Predicting Model Failure using Saliency Maps in Autonomous Driving Systems} 



\icmlsetsymbol{equal}{*}

\begin{icmlauthorlist}
\icmlauthor{Sina Mohseni}{to}
\icmlauthor{Akshay Jagadeesh}{to}
\icmlauthor{Zhangyang Wang}{to}
\end{icmlauthorlist}

\icmlaffiliation{to}{Department of Computer Science and Engineering, Texas A\&M University, College Station}

\icmlcorrespondingauthor{Sina Mohseni}{sina.mohseni@tamu.edu}

\icmlkeywords{Machine Learning, Failure Prediction, Interpretability}

\vskip 0.3in
]



\printAffiliationsAndNotice{}  


\begin{abstract}
   While machine learning systems show high success rate in many complex tasks, research shows they can also fail in very unexpected situations. 
   Rise of machine learning products in safety-critical industries cause an increase in attention in evaluating model robustness and estimating failure probability in machine learning systems.
   In this work, we propose a design to train a student model -- a failure predictor -- to predict the main model's error for input instances based on their saliency map.
   We implement and review the preliminary results of our failure predictor model on an autonomous vehicle steering control system as an example of safety-critical applications.
\end{abstract}
\section{Introduction}
How can we evaluate machine learning systems for safety critical domains such as autonomous driving?
While machine learning systems show high success rates in many complex tasks such as speech recognition and image captioning, research shows they can also fail in unexpected situations~\cite{goodfellow2014explaining} and even cause fatality in critical applications~\cite{kohli2019enabling}.
The goal of this project is to predict model failure in safety-critical applications, such as automated vehicles, where failure costs are extremely high compared to other commercial machine learning products, like targeted advertisement algorithms.
Another motivation of this research is to further improve machine learning products for real-world applications. 
An important example of industrial level requirement for machine learning products is providing fail-safe conditions in functional safety standards like ISO26262 in the auto industry.

Recently, there has been increasing attention in evaluating model robustness and failure estimation in machine learning research~\cite{kendall2017uncertainties}.
Although model robustness has multiple and different definitions in the literature, it boils down to validating non-uniform model behavior.
Researchers study the roots of model behavior in different topics such as adversarial attacks, architecture and training process inefficiency, and training data incompleteness.
New model architectures like Bayesian Deep Learning~\cite{gal2016uncertainty} opened new doors to perform failure estimation via epistemic uncertainty analysis~\cite{uesato2018rigorous}, but with the growing popularity of pre-trained machine learning models and APIs with possibly biased or insufficient training data, it is important to analyze the model for blind spots.
 
In this paper, we propose a new design to train a \textit{failure predictor} student model for predicting the main model's error for every input instance.
We implement and evaluate our method on a common machine learning solution in self-driving car applications as an important safety-critical domain.
We use PilotNet~\cite{bojarski2016end}, as a case study, which is a driving model to control the steering wheel. 
We then train a failure predictor model using saliency maps from the main model to predict the steering wheel angle error.
We evaluate our failure predictor model based on prediction error and the driving safety gained by this system.

\begin{figure*}[t!]
\begin{center}
\includegraphics[width=0.9\linewidth]{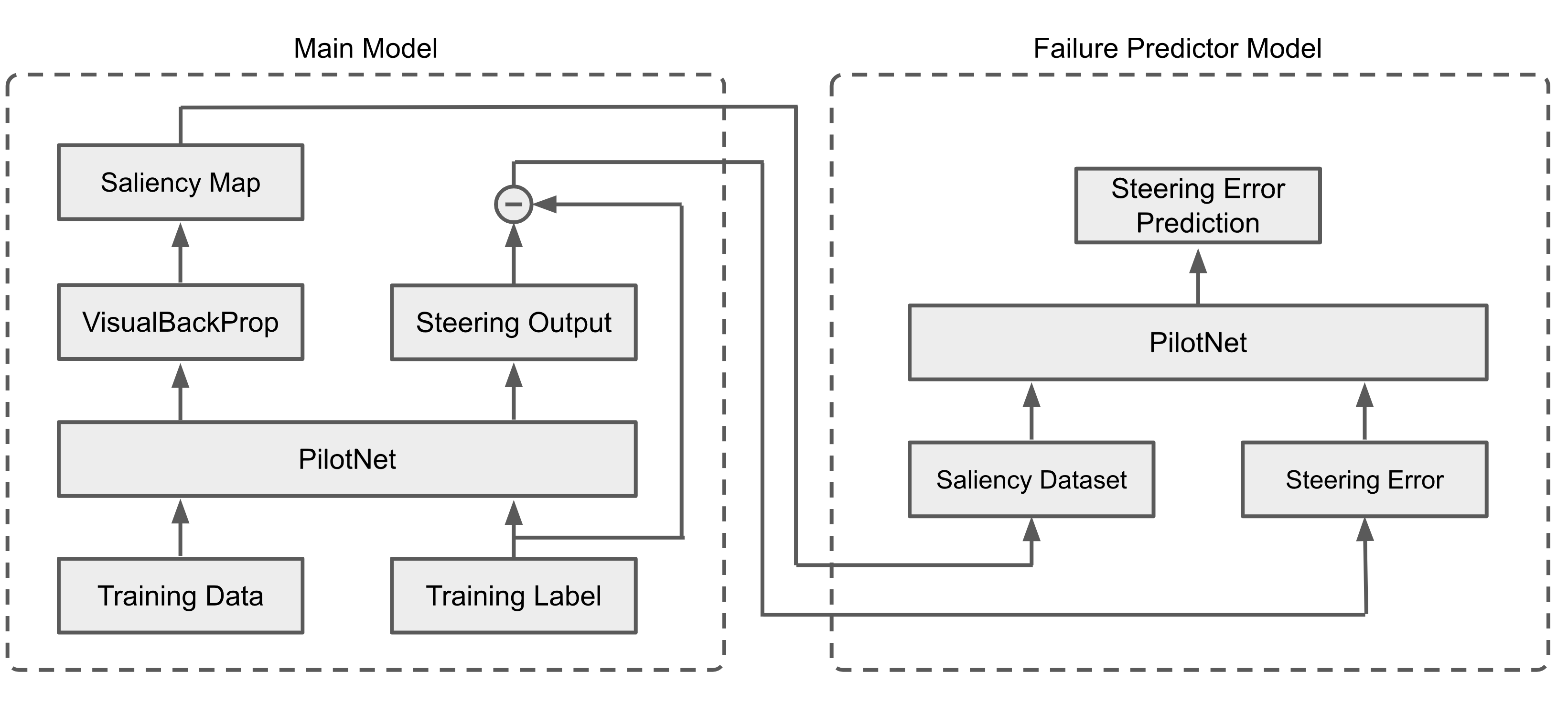}  %
\end{center}
   \caption{Overview of our method to train the failure predictor model.}
\label{fig:pipeline}
\end{figure*}

\section{Background}
Here we briefly review two main directions for identifying model failure types including predicting failure instances and improving model resilience.

Exploring model blind spots due to random biases in the training data is an active line of research that is studying model robustness.
For example, in reinforcement learning architectures, Uesato et al.~\cite{uesato2018rigorous} propose an adversarial approach to search for scenarios where an agent could fail. 
Similarly, to find the non-uniform error behavior of the system in image captioning tasks, Nushi et al.~\cite{nushi2018towards} propose a clustering method to observe and model the failure conditions with decision trees.
Others also investigated different search algorithms (e.g., ~\cite{bansal2018coverage, lakkaraju2017identifying}) and crowdsourcing approaches (e.g., ~\cite{attenberg2015beat}) to identify model failures assuming there is no access to the training data.

Furthermore, recent machine learning evaluations show how a model's over reliance on context could also cause unintended context dependencies which may result in the model hallucination of objects that are not present in the image~\cite{shetty2018not}.  
In another paper, Rosenfeld et al.~\cite{rosenfeld2018elephant} studied objects location dependency in image recognition models.

Adversarial examples are another class of artifacts that question model robustness by causing machine learning models to misclassify inputs with high confidence~\cite{goodfellow2014explaining}.
Since new attack techniques are being designed by the machine learning community, new defense mechanisms have been also invented to increase model robustness~\cite{biggio2018wild}.
Related to our work, Papernot et al.~\cite{papernot2016distillation} show model distillation can improve model robustness against adversarial attacks.
They define their robustness metric and, in a set of evaluation experiments, show how the distilled student model is more robust to adversary attacks.
Recently, Fong and Vedaldi~\cite{fong2017interpretable} show the application of salient maps in detecting adversarial attacks. 
They train an AlexNet model to distinguish between clean and adversarial images (using a one-step iterative method from~\cite{kurakin2016adversarial}) and achieved high accuracy.
Similarly, Zhang et al.~\cite{zhang2018detecting} designed an adversarial example detector by concatenating input images with the saliency maps to train a binary adversary detector.
However, rigorous evaluations of defense methods~\cite{Carlini:2017:AEE:3128572.3140444} demonstrate certain limitations of current defenses when dealing with complex attacks.

Our work differs with the current machine learning research as we focus on the model's error (or failure) probability estimation in which a student model is trained by a weak teacher.
To further digest and pass the useful learning representations from the teacher to the student model (failure predictor model), we create a saliency map training set for the student model.
\section{Methodology}

We propose a transfer learning approach in which we train a \textit{failure predictor} model to predict the main model's error at each prediction.
We follow \cite{papernot2016distillation} to transfer knowledge from the initial model to a student model with the same architecture. 
Similar to the approach in~\cite{fong2017interpretable}, we create a saliency training set to train the failure predictor model. 
The saliency maps are generated from the main model and we utilize the VisualBackProp~\cite{bojarski2016visualbackprop} technique to generate saliency maps.
The choice of VisualBackProp technique is because of the fact that this technique is originally designed and tested for the PilotNet model and images including roads and vehicles.
The training labels for the failure prediction model are the steering wheel angle (SWA) prediction errors of the initial model for each frame.

\textit{New Target = PilotNet (SWA Prediction) - Baseline (SWA)}

Figure~\ref{fig:pipeline} shows an overview of our pipeline for training the failure predictor model along with the initial model.
Since a highly accurate teacher model cannot transfer its weaknesses and blind spots to the failure predictor model, similar to Uesato et al.~\cite{uesato2018rigorous}, we use the same architecture but a weakly trained teacher model to create our saliency map training set.
In our implementation and experiments, we also had to transfer the convolutional layers from the teacher model to the student model to start the training. 
This may be due to the less informative nature of saliency maps compared to the original images.

\section{Implementation}

In this section, we review the implementation details of our failure prediction technique on the PilotNet algorithm as an example of a steering wheel angle prediction algorithm.

\begin{figure}[t!]
\begin{center}
\includegraphics[width=0.9\linewidth]{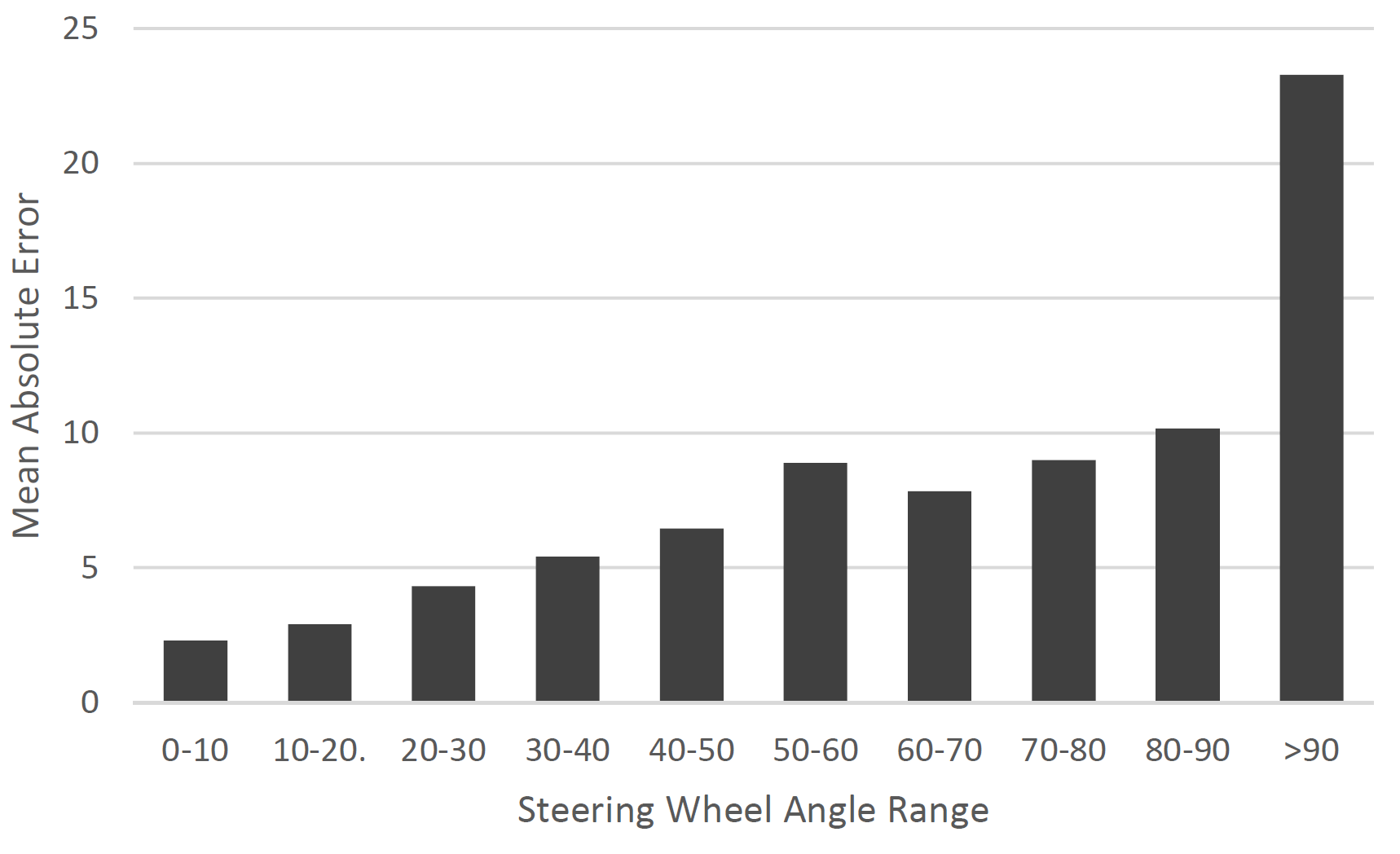}  
\end{center}
   \caption{Trained PilotNet model performance across driving conditions. The performance graph suggests the model prediction error increases with the road's degree of curvature.}
\label{fig:pilotNet-MAE}
\end{figure}

\subsection{PilotNet Model}

We borrow the PilotNet architecture form from Bojarski et. al~\cite{bojarski2016end} as a simple network for a SWA prediction.
PilotNet includes five convolutional layers followed by three fully connected layers and no softmax layer.
PilotNet's output is the prediction SWA which is the inverse turning radius.
Our training data was limited to 54,000 samples from one front camera and no other driving or navigation related data.
Input camera images were converted to grayscale, cropped to the top-half of the image, and resized to 102 by 364 pixels.
We have not used any other pre-processing or data augmentation techniques in this implementation.
We used drop-out layers for the fully connected layers and Adam optimizer~\cite{kingma2014adam} for the model training.
Figure~\ref{fig:pilotNet-MAE} shows the Mean Absolute Error (MAE) of our trained model for ranges of SWA.
These SWA prediction results show PilotNet has a weak performance in predicting SWA at intersections and sharp turns.
This may be due to the fact that no navigation data is used in PilotNet training. 

\subsection{Saliency Map}

\begin{figure}[t!]
\begin{center}
\includegraphics[width=0.9\linewidth]{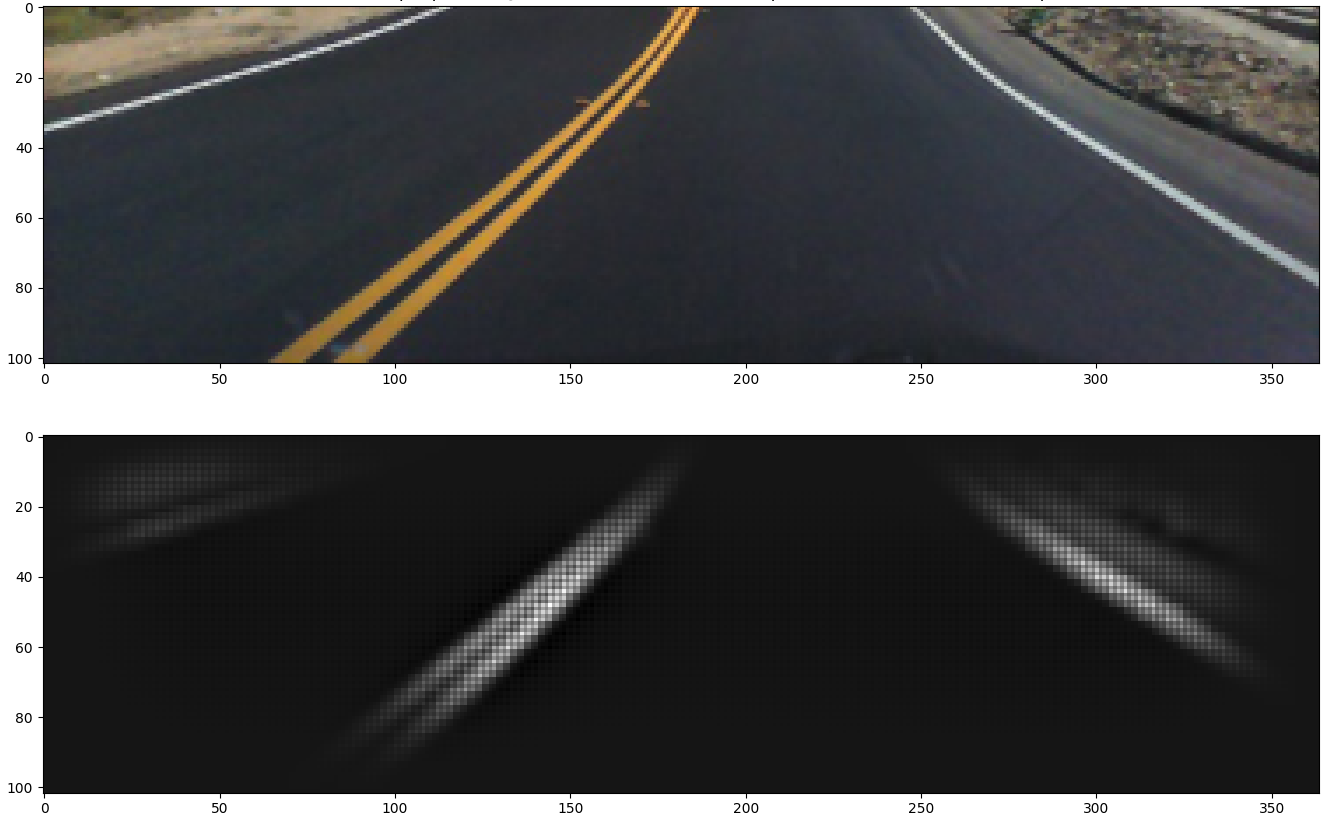}  %
\end{center}
   \caption{An example of saliency maps of the input images. \textbf{Top:} Input camera image \textbf{Bottom:} Generated saliency map using VisualBackProp~\cite{bojarski2016visualbackprop} technique.}
\label{fig:saliency}
\end{figure}

In order to generate the saliency maps, we use VisualBackProp technique from Bojarski et.al~\cite{bojarski2016visualbackprop}.
VisualBackProp layer-wise relevance propagation technique involves a series of deconvolution and point-wise multiplication of averaged feature maps in each layer to upscale the previous layer.
Figure~\ref{fig:saliency} shows an example of a saliency map of the input camera image using VisualBackProp technique.
We then generate saliency maps for each frame to create a saliency map training set to train the failure predictor model. To optimize our saliency training set for maximum of failure cases but limited to the dataset distribution, we use a weakly trained main model to create the saliency maps.
We also calculate the SWA error and save as the training target for the failure predictor model.
This is done through testing the trained PilotNet model and calculating the difference between the predicted SWA and baseline SWA for each input image.

\subsection{Failure Predictor Model}

We choose the the same network architecture as PilotNet model for our failure predictor model due to their task similarities.
However, it is trained with the saliency maps training set instead of camera images.  
The saliency map inputs are in grayscale with the same size as the camera images.
Also, different from the main model, the training target for the failure predictor model is the SWA prediction error of the main model at each frame.
We also transfer pre-trained convolution layers from the main model to start the training of our failure predictor model.
This improves the training of the failure predictor model by picking the training up from the main model. 
We optimized the failure predictor model in 30 epochs with batch size of 128 using Adam optimizer and learning rate of 1e-5.
We review the performance of this model in the next section.

\section{Evaluation}

We use two evaluation methods to measure our method's performance in predicting model failure.
We first calculate and report the failure predictor model's MAE and then compare our results with Hecker et al.~\cite{hecker2018failure} 
in terms of increasing driving safety in a human-machine collaborative driving experience.
We also compare the failure predictor model performance when trained with saliency maps versus camera images.

\subsection{Evaluating Failure Prediction}

\begin{figure}[t!]
\begin{center}
\includegraphics[width=0.9\linewidth]{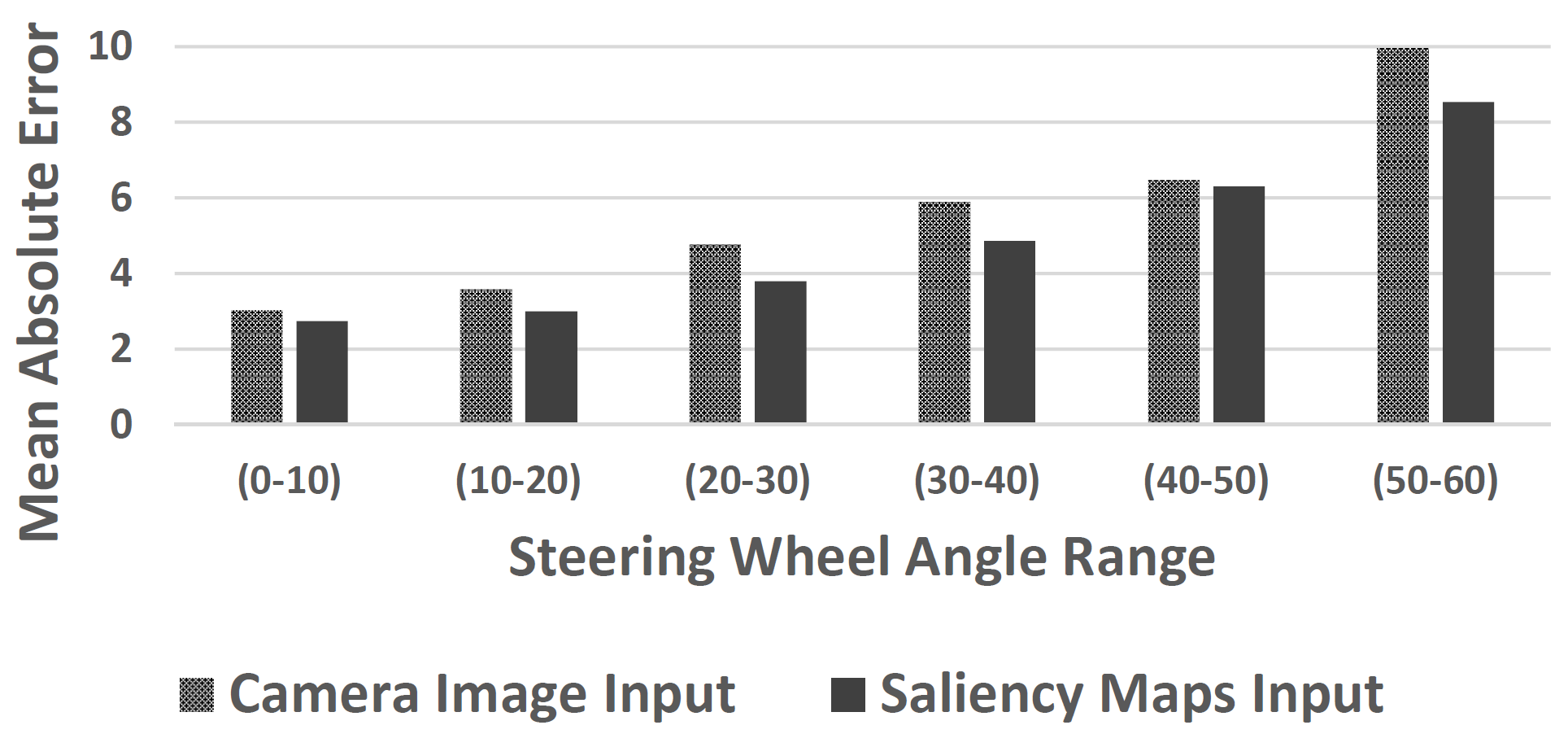}  
\end{center}
    \caption{Comparison of the failure Prediction Model performance when trained with saliency maps versus camera input images. Similar trend observed in higher SWA ranges.}
\label{fig:student-MAE}
\end{figure}

We first evaluate the performance of our failure prediction model using the MAE of predicting the main model's (PilotNet) error. 
The predictor model's MAE is the mean absolute deviation of each frame's ``predicted error'' from ``baseline error''.
Figure~\ref{fig:student-MAE} presents the MAE of the failure predictor model across different ranges of SWA. 
Similar to the observation of PilotNet's accuracy (see figure~\ref{fig:pilotNet-MAE}), failure predictor model also performs better in slight turns and curves such as on highways.
We also present a comparison between error prediction performance when the student model is trained with image inputs versus saliency maps.
This comparison indicates that saliency inputs have enough information to perform a better failure prediction.

\subsection{Evaluating Safety Gain}

In our second evaluation, we choose an error threshold as the unsafe condition to be predicted by the failure prediction model.
In this evaluation case, the failure prediction model serves as a safety mechanism to reduce the risk of collisions by alerting the driver to take over.
Similar to~\cite{hecker2018failure}, we define five degrees as the SWA prediction error threshold for the unsafe conditions. 
We then calculate the true positive rate in predicting failure instances based on this threshold. 

The true positive rate shows the overall safety gained by alerting the driver (in case of model failure) and reducing the risk of collision. 
Table~\ref{tab:safety} shows the safety gained evaluation result across different SWA ranges.
Comparing the image and saliency map inputs shows saliency maps superiority in detecting the model failure and increasing driving safety.

\begin{table}
\centering
\begin{tabular}{||c | c c||}
 \hline
 \multirow{ 2}{*}{SWA Range}  & \multicolumn{2}{c||}{Unsafe Condition Threshold} \\
  \cline{2-3}
  & Saliency Map & Camera Image\\
 \hline
 (0\textdegree-30\textdegree) SWA & 42\% & 18\% \\
 (30\textdegree-60\textdegree) SWA & 56\% & 51\% \\
 (60\textdegree-90\textdegree) SWA & 68\% & 68\% \\
 \hline
 
\end{tabular}
\caption{Safety gained with our failure predictor model across different SWA ranges. Higher safety gain in higher SWA ranges could be a result of the main model's poor performance on those conditions.}
\label{tab:safety}
\end{table}


\section{Discussion and Future Works}
Saliency maps are visualizations of what a model has learned relevant to each individual input.
Although this interpretability method is primarily being used to understand and debug the machine learning models, we take advantage of saliency maps to evaluate a model's safety and failure prediction. 
Different methods have been proposed to evaluate (e.g., ~\cite{samek2017evaluating} and ~\cite{mohseni2018human}) and perform sanity checks (e.g., ~\cite{adebayo2018sanity}) on saliency methods. 
Different from others, we are aiming to maintain the safety of the machine learning software by predicting its reliability through a white-box approach.
In our future work, we plan to compare how different saliency methods would affect failure prediction performance. 

Comparison of our preliminary evaluation results with~\cite{hecker2018failure} suggests saliency maps' higher performance in predicting model failure, Table~\ref{tab:safety}. 
However, We could not compare our results with the uncertainty-based failure prediction method proposed in~\cite{michelmore2018evaluating} due to their very limited training and evaluation experiment conditions.
To better exploit saliency maps in predicting model reliability, in our future work, we plan to increase training data size and augment different noises such as rain and fog for a better evaluation experiment. 
Our experiments show evaluation criteria in autonomous driving algorithms are very diverse, and the use of driving simulators help to cover more safety measures such as vehicle deviation from baseline trajectory and collision prediction. 
\section{Conclusion}

In this paper, we proposed a design for utilizing saliency maps to predict model failure in a safety-critical application.
We compared our method with similar papers in the autonomous vehicles safety field.
Our preliminary experiments and evaluation shows saliency maps out perform raw images in predicting model failure and are possible candidates to predict model failure in case of adverse conditions.


\bibliography{biblograph}

\begin{thebibliography}{25}
\providecommand{\natexlab}[1]{#1}
\providecommand{\url}[1]{\texttt{#1}}
\expandafter\ifx\csname urlstyle\endcsname\relax
  \providecommand{\doi}[1]{doi: #1}\else
  \providecommand{\doi}{doi: \begingroup \urlstyle{rm}\Url}\fi

\bibitem[Adebayo et~al.(2018)Adebayo, Gilmer, Muelly, Goodfellow, Hardt, and
  Kim]{adebayo2018sanity}
Adebayo, J., Gilmer, J., Muelly, M., Goodfellow, I., Hardt, M., and Kim, B.
\newblock Sanity checks for saliency maps.
\newblock In \emph{Advances in Neural Information Processing Systems}, pp.\
  9505--9515, 2018.

\bibitem[Attenberg et~al.(2015)Attenberg, Ipeirotis, and
  Provost]{attenberg2015beat}
Attenberg, J., Ipeirotis, P., and Provost, F.
\newblock Beat the machine: Challenging humans to find a predictive model's
  “unknown unknowns”.
\newblock \emph{Journal of Data and Information Quality (JDIQ)}, 6\penalty0
  (1):\penalty0 1, 2015.

\bibitem[Bansal \& Weld(2018)Bansal and Weld]{bansal2018coverage}
Bansal, G. and Weld, D.~S.
\newblock A coverage-based utility model for identifying unknown unknowns.
\newblock In \emph{Thirty-Second AAAI Conference on Artificial Intelligence},
  2018.

\bibitem[Biggio \& Roli(2018)Biggio and Roli]{biggio2018wild}
Biggio, B. and Roli, F.
\newblock Wild patterns: Ten years after the rise of adversarial machine
  learning.
\newblock \emph{Pattern Recognition}, 84:\penalty0 317--331, 2018.

\bibitem[Bojarski et~al.(2016{\natexlab{a}})Bojarski, Choromanska, Choromanski,
  Firner, Jackel, Muller, and Zieba]{bojarski2016visualbackprop}
Bojarski, M., Choromanska, A., Choromanski, K., Firner, B., Jackel, L., Muller,
  U., and Zieba, K.
\newblock Visualbackprop: visualizing cnns for autonomous driving.
\newblock \emph{arXiv preprint arXiv:1611.05418}, 2, 2016{\natexlab{a}}.

\bibitem[Bojarski et~al.(2016{\natexlab{b}})Bojarski, Del~Testa, Dworakowski,
  Firner, Flepp, Goyal, Jackel, Monfort, Muller, Zhang,
  et~al.]{bojarski2016end}
Bojarski, M., Del~Testa, D., Dworakowski, D., Firner, B., Flepp, B., Goyal, P.,
  Jackel, L.~D., Monfort, M., Muller, U., Zhang, J., et~al.
\newblock End to end learning for self-driving cars.
\newblock \emph{arXiv preprint arXiv:1604.07316}, 2016{\natexlab{b}}.

\bibitem[Carlini \& Wagner(2017)Carlini and
  Wagner]{Carlini:2017:AEE:3128572.3140444}
Carlini, N. and Wagner, D.
\newblock Adversarial examples are not easily detected: Bypassing ten detection
  methods.
\newblock In \emph{Proceedings of the 10th ACM Workshop on Artificial
  Intelligence and Security}, AISec '17, pp.\  3--14, New York, NY, USA, 2017.
  ACM.
\newblock ISBN 978-1-4503-5202-4.
\newblock \doi{10.1145/3128572.3140444}.
\newblock URL \url{http://doi.acm.org/10.1145/3128572.3140444}.

\bibitem[Fong \& Vedaldi(2017)Fong and Vedaldi]{fong2017interpretable}
Fong, R.~C. and Vedaldi, A.
\newblock Interpretable explanations of black boxes by meaningful perturbation.
\newblock In \emph{Proceedings of the IEEE International Conference on Computer
  Vision}, pp.\  3429--3437, 2017.

\bibitem[Gal(2016)]{gal2016uncertainty}
Gal, Y.
\newblock \emph{Uncertainty in deep learning}.
\newblock PhD thesis, PhD thesis, University of Cambridge, 2016.

\bibitem[Goodfellow et~al.(2014)Goodfellow, Shlens, and
  Szegedy]{goodfellow2014explaining}
Goodfellow, I.~J., Shlens, J., and Szegedy, C.
\newblock Explaining and harnessing adversarial examples.
\newblock \emph{arXiv preprint arXiv:1412.6572}, 2014.

\bibitem[Hecker et~al.(2018)Hecker, Dai, and Van~Gool]{hecker2018failure}
Hecker, S., Dai, D., and Van~Gool, L.
\newblock Failure prediction for autonomous driving.
\newblock In \emph{2018 IEEE Intelligent Vehicles Symposium (IV)}, pp.\
  1792--1799. IEEE, 2018.

\bibitem[Kendall \& Gal(2017)Kendall and Gal]{kendall2017uncertainties}
Kendall, A. and Gal, Y.
\newblock What uncertainties do we need in bayesian deep learning for computer
  vision?
\newblock In \emph{Advances in neural information processing systems}, pp.\
  5574--5584, 2017.

\bibitem[Kingma \& Ba(2014)Kingma and Ba]{kingma2014adam}
Kingma, D.~P. and Ba, J.
\newblock Adam: A method for stochastic optimization.
\newblock \emph{arXiv preprint arXiv:1412.6980}, 2014.

\bibitem[Kohli \& Chadha(2019)Kohli and Chadha]{kohli2019enabling}
Kohli, P. and Chadha, A.
\newblock Enabling pedestrian safety using computer vision techniques: A case
  study of the 2018 uber inc. self-driving car crash.
\newblock In \emph{Future of Information and Communication Conference}, pp.\
  261--279. Springer, 2019.

\bibitem[Kurakin et~al.(2016)Kurakin, Goodfellow, and
  Bengio]{kurakin2016adversarial}
Kurakin, A., Goodfellow, I., and Bengio, S.
\newblock Adversarial examples in the physical world.
\newblock \emph{arXiv preprint arXiv:1607.02533}, 2016.

\bibitem[Lakkaraju et~al.(2017)Lakkaraju, Kamar, Caruana, and
  Horvitz]{lakkaraju2017identifying}
Lakkaraju, H., Kamar, E., Caruana, R., and Horvitz, E.
\newblock Identifying unknown unknowns in the open world: Representations and
  policies for guided exploration.
\newblock In \emph{Thirty-First AAAI Conference on Artificial Intelligence},
  2017.

\bibitem[Michelmore et~al.(2018)Michelmore, Kwiatkowska, and
  Gal]{michelmore2018evaluating}
Michelmore, R., Kwiatkowska, M., and Gal, Y.
\newblock Evaluating uncertainty quantification in end-to-end autonomous
  driving control.
\newblock \emph{arXiv preprint arXiv:1811.06817}, 2018.

\bibitem[Mohseni \& Ragan(2018)Mohseni and Ragan]{mohseni2018human}
Mohseni, S. and Ragan, E.~D.
\newblock A human-grounded evaluation benchmark for local explanations of
  machine learning.
\newblock \emph{arXiv preprint arXiv:1801.05075}, 2018.

\bibitem[Nushi et~al.(2018)Nushi, Kamar, and Horvitz]{nushi2018towards}
Nushi, B., Kamar, E., and Horvitz, E.
\newblock Towards accountable ai: Hybrid human-machine analyses for
  characterizing system failure.
\newblock In \emph{Sixth AAAI Conference on Human Computation and
  Crowdsourcing}, 2018.

\bibitem[Papernot et~al.(2016)Papernot, McDaniel, Wu, Jha, and
  Swami]{papernot2016distillation}
Papernot, N., McDaniel, P., Wu, X., Jha, S., and Swami, A.
\newblock Distillation as a defense to adversarial perturbations against deep
  neural networks.
\newblock In \emph{2016 IEEE Symposium on Security and Privacy (SP)}, pp.\
  582--597. IEEE, 2016.

\bibitem[Rosenfeld et~al.(2018)Rosenfeld, Zemel, and
  Tsotsos]{rosenfeld2018elephant}
Rosenfeld, A., Zemel, R., and Tsotsos, J.~K.
\newblock The elephant in the room.
\newblock \emph{arXiv preprint arXiv:1808.03305}, 2018.

\bibitem[Samek et~al.(2017)Samek, Binder, Montavon, Lapuschkin, and
  M{\"u}ller]{samek2017evaluating}
Samek, W., Binder, A., Montavon, G., Lapuschkin, S., and M{\"u}ller, K.-R.
\newblock Evaluating the visualization of what a deep neural network has
  learned.
\newblock \emph{IEEE transactions on neural networks and learning systems},
  28\penalty0 (11):\penalty0 2660--2673, 2017.

\bibitem[Shetty et~al.(2018)Shetty, Schiele, and Fritz]{shetty2018not}
Shetty, R., Schiele, B., and Fritz, M.
\newblock Not using the car to see the sidewalk: Quantifying and controlling
  the effects of context in classification and segmentation.
\newblock \emph{arXiv preprint arXiv:1812.06707}, 2018.

\bibitem[Uesato et~al.(2018)Uesato, Kumar, Szepesvari, Erez, Ruderman,
  Anderson, Heess, Kohli, et~al.]{uesato2018rigorous}
Uesato, J., Kumar, A., Szepesvari, C., Erez, T., Ruderman, A., Anderson, K.,
  Heess, N., Kohli, P., et~al.
\newblock Rigorous agent evaluation: An adversarial approach to uncover
  catastrophic failures.
\newblock \emph{arXiv preprint arXiv:1812.01647}, 2018.

\bibitem[Zhang et~al.(2018)Zhang, Ye, Wang, and Yang]{zhang2018detecting}
Zhang, C., Ye, Z., Wang, Y., and Yang, Z.
\newblock Detecting adversarial perturbations with saliency.
\newblock In \emph{2018 IEEE 3rd International Conference on Signal and Image
  Processing (ICSIP)}, pp.\  271--275. IEEE, 2018.

\end{thebibliography}
\bibliographystyle{icml2019}






\end{document}